\documentclass{llncs}
\usepackage{makeidx}  
\usepackage{llncsdoc}
\usepackage{mathtools}
\usepackage{amsfonts}
\usepackage{booktabs}
\usepackage{multirow}
\usepackage{array}
\usepackage{float}
\usepackage{sidecap}
\usepackage{wrapfig}
\usepackage{afterpage}
\usepackage{caption}
\usepackage{graphicx}
\usepackage{subcaption}
\usepackage{booktabs}
\usepackage{multirow}
\usepackage{float}
\usepackage{placeins}
\usepackage{array}
\usepackage{wrapfig}
\usepackage[acronym]{glossaries}

\newcolumntype{C}[1]{>{\centering\arraybackslash}p{#1}}

\newacronym{mri}{MRI}{Magnetic Resonance Imaging}
\newacronym{mr}{MR}{Magnetic Resonance}
\newacronym{pet}{PET}{Positron Emission Tomography}
\newacronym{gan}{GAN}{Generative Adversarial Network}
\newacronym{ml}{ML}{Machine Learning}
\newacronym{ct}{CT}{Computed Tomography}
\newacronym{wmhpvo}{WMH$_{\mathrm{pvo}}$}{White Matter Hyperintensity of Presumed Vascular Origin}
\newacronym{flair}{FLAIR}{Fluid-Attenuated Inversion Recovery}
\newacronym{svd}{SVD}{Small Vessel Disease}
\newacronym{wmh}{WMH}{White Matter Hyperintensity}
\newacronym{wm}{WM}{White Matter}
\newacronym{gm}{GM}{Grey Matter}
\newacronym{csf}{CSF}{Cerebrospinal Fluid}
\newacronym{3d}{3D}{3-dimensional}
\newacronym{2d}{2D}{2-dimensional}
\newacronym{4d}{4D}{4-dimensional}
\newacronym{auc}{AUC}{Area Under the Curve}
\newacronym{roi}{ROI}{Region of Interest}
\newacronym{ms}{MS}{Multiple Sclerosis}
\newacronym{cnn}{CNN}{Convolutional Neural Network}
\newacronym{malpem}{MALPEM}{Multi-Atlas-Label Propagation with Expectation-Maximisation based refinement}
\newacronym{pggan}{PGGAN}{Progressive Growing of GANs}
\newacronym{wgan}{WGAN}{Wasserstein Generative Adversarial Network}
\newacronym{dcgan}{DCGAN}{Deep Convolutional Generative Adversarial Network}
\newacronym{mse}{MSE}{Mean Squared Error}
\newacronym{dsc}{DSC}{Dice Similarity Coefficient}
\newacronym{mni}{MNI}{Montreal Neurological Institute}
\newacronym{ffd}{FFD}{Free Form Deformation}

\newcolumntype{P}[1]{>{\centering\arraybackslash}p{#1}}
\newcolumntype{M}[1]{>{\centering\arraybackslash}m{#1}}

\begin{document}
\pagestyle{headings}  
\pagestyle{empty}
\addtocmark{GAN Augmentation: Augmenting Training Data using Generative Adversarial Networks} 
\mainmatter              
\title{GAN Augmentation: Augmenting Training Data using Generative Adversarial Networks}
\titlerunning{GAN Augmentation}  
%
\author{*****}
\institute{**}
\author{Christopher Bowles\inst{1} \and Liang Chen\inst{1} \and Ricardo Guerrero\inst{1,6}$^*$ \and Paul Bentley\inst{2} \and Roger Gunn\inst{2,3}\and Alexander Hammers\inst{4} \and David Alexander Dickie\inst{5,7} \and Maria  Vald\'{e}s Hern\'{a}ndez\inst{5} \and Joanna Wardlaw\inst{5} \and  Daniel Rueckert\inst{1}}


%
%
%
\institute{Department of Computing, Imperial College London, UK \and Department of Medicine, Imperial College London, UK \and Imanova Ltd., London, UK \and PET Centre, King’s College London, UK  \and Department of Neuroimaging Sciences, University of Edinburgh, UK \and Samsung AI Research Centre (SAIC), Cambridge, UK \and Institute of Cardiovascular and Medical Sciences, University of Glasgow, UK}

\maketitle              
\section*{Abstract}
One of the biggest issues facing the use of machine learning in medical imaging is the lack of availability of large, labelled datasets. The annotation of medical images is not only expensive and time consuming but also highly dependent on the availability of expert observers. The limited amount of training data can inhibit the performance of supervised machine learning algorithms which often need very large quantities of data on which to train to avoid overfitting. So far, much effort has been directed at extracting as much information as possible from what data is available. \Glspl{gan} offer a novel way to unlock additional information from a dataset by generating synthetic samples with the appearance of real images. This paper demonstrates the feasibility of introducing \gls{gan} derived synthetic data to the training datasets in two brain segmentation tasks, leading to improvements in \gls{dsc} of between 1 and 5 percentage points under different conditions, with the strongest effects seen fewer than ten training image stacks are available. \let\thefootnote\relax\footnote{$^*$ All work done while at Imperial College London}

\section{Introduction}

Data augmentation is commonly used by many deep learning approaches in the presence of limited training data. Increasing the number of training examples through the rotation, reflection, cropping, translation and scaling of existing images is common practice during the training of learning algorithms, allowing for the number of samples in a dataset to be increased by factors of thousands~\cite{krizhevsky2012imagenet}. Populating the training data with realistic, if synthetic, data in this way can significantly reduce overfitting and thus not only improve the accuracy but also the generalisation ability of deep learning approaches. This is of particular importance in \glspl{cnn} which cannot easily learn rotationally invariant features unless there are sufficient examples at different rotations in the training data. This paper investigates using a \gls{gan} to model the underlying distribution of training data to allow for additional synthetic data to be sampled and used to augment the real training data.

First proposed in~\cite{Goodfellow2014}, \glspl{gan} are a class of neural networks which learn to generate synthetic samples with the same characteristics as a given training distribution. In the case of images, this involves learning to produce images (via a generator) which are visually so similar to a set of real images that an adversary (the discriminator) cannot detect them. The original formulation has been built on to address problems such as training stability~\cite{Radford2015}, low resolution~\cite{nvidia}, and the absence of a true image quality based loss function~\cite{Arjovsky2017}, and applied to tasks such as super resolution~\cite{Ledig2016}, reconstructing images from a minimal data~\cite{Yeh2016} and anomaly detection~\cite{schlegl2017unsupervised}.

Various methods for using \glspl{gan} to expand training datasets have been recently proposed. In~\cite{CVPR}, the authors use an adversarial network to improve the quality of simulated images, and use these for further training. In~\cite{Antoniou}, the authors train a conditional \gls{gan} on unlabelled data to generate alternative versions of a given real image, and in~\cite{Zhu2017}, the authors use a similar \gls{gan} to impose emotions on neutral faces to expand underrepresented classes. However, the use of non-conditional \glspl{gan} to augment training data directly as a preprocessing step with no additional data has only very recently been explored~\cite{Frid-Adar2018,SPIE}, with promising results in medical image classification tasks.

\subsection{Motivation}


Imaging features can be divided into two categories, measuring either \textit{pertinent} or \textit{non-pertinent} variance. Pertinent features are those which are important to whatever information the user wishes to extract. In medical imaging, these are features such as the size, shape, intensity and location of key components such as organs or lesions. Non-pertinent features are those which vary between images but are unrelated to the information the user wishes to extract. Examples of these are global intensity differences, position within the image field of view and appearance of unrelated anatomy. Exactly which features are pertinent or non-pertinent will depend on the application, and may not be known a-priori. 

A lot of non-pertinent variance can easily be removed from a dataset. Common methods include intensity normalisation, cropping, and registration to a standard space. These processes substantially simplify the data distribution, and importantly, can be applied to test instances. Keeping too much non-pertinent variance can not only occlude the diagnostically important information, but also lead to overfitting, especially in the small datasets often used in medical imaging.


Data augmentation is an alternative to removing non-pertinent variance. One of the goals of data augmentation is to populate the data with a large amount of synthetic data in the directions of these non-pertinent sources of variance. The aim of this is to reduce this variance to noise, removing any coincidental correlation with labels and preventing its use as a discriminative feature. 

As noted in~\cite{deform}, there is a tendency within medical imaging to remove non-pertinent variance rather than use augmentation. This due to both the ease with which much of the non-pertinent variance can be removed, and the lack of suitable augmentation procedures for many sources of non-pertinent variance. This is reflected in~\cite{CRF}, where the authors choose to only employ reflection and intensity augmentation for brain lesion segmentation, with even the latter omitted when using larger datasets. On the other hand, in~\cite{UNet} the authors benefit from extensive augmentation in their application of microscopy images, particularly through random elastic deformations. This demonstrates how careful consideration of the application will inform which types of augmentation are appropriate. While random elastic deformations may be an appropriate model for microscopy images, in which the objects of interest (cells) are generally fluid and unconstrained, applying the same procedure to brain images could lead to certain anatomical constraints such as symmetry, rigidity, and structure being disregarded. In addition, some sources of non-pertinent variance can be neither removed nor augmented by traditional means. For example, patient specific variation in non-relevant anatomy, where it may not be possible to remove this anatomy through cropping, or to define an accurate enough model to augment this variance with realistic cases.


\Glspl{gan} offer a potentially valuable addition to the arsenal of augmentation techniques which are currently available. One of the main potential advantages of \glspl{gan} is that they take many decisions away from the user, in much the same way that deep learning removes the need for ``hand crafted'' features. An ideal \gls{gan} will transform the discrete distribution of training samples into a continuous distribution, thereby simultaneously applying augmentation to each source of variance within the dataset. For example, given a sufficient number of training examples at different orientations, a \gls{gan} will learn to produce examples at any orientation, replicating the effects of applying rotation augmentation. While orientation is a source of variance which can easily be augmented or removed using traditional methods, consider instead a more challenging source of variance such as ventricle size in brain imaging. Again, given a sufficient number of training examples of patients with different discrete ventricle sizes, a trained \gls{gan} will be able to produce examples along the continuum of all sizes. To perform the same kind of augmentation using deformations would involve a complex model of realistic ventricle size, shape and impact on the surrounding anatomy. By simultaneously learning the distribution of all sources of variance, the \gls{gan} infers this model directly from the available data. 

One potential limitation of using \glspl{gan} for augmentation is their ability to generate images with a high enough quality. While improvements have been made, \glspl{gan} cannot be relied upon to produce images with perfect fidelity. 
This is not a problem for traditional augmentation procedures which do not significantly degrade the images. However, both~\cite{Fischer2015} and~\cite{Richter2016} demonstrate that complete realism is not necessary to improve results with synthetic data. Whether the advantage of additional data is outweighed by the disadvantage of lower quality images is one of the questions we address in this paper.

\subsection{Contribution}

The results reported in~\cite{Frid-Adar2018,SPIE} suggest that \glspl{gan} can have a significant benefit when used for data augmentation in some classification tasks. In this paper we thoroughly investigate this use of \glspl{gan} in different domains for the purpose of medical image segmentation. An in depth investigation into the effects of \gls{gan} augmentation is first carried out on a complex multi-class \gls{ct} \gls{csf} segmentation task using two segmentation architectures. By choosing not to co-register the images in this dataset, we are able to examine how \gls{gan} augmentation compares and interacts with rotation augmentation. The transferability of the method is then evaluated by applying it to a second dataset of \gls{flair} \gls{mr} images for the purpose of single-class \gls{wmh} segmentation. This is a well studied problem, and poses challenges typical to medical image segmentation tasks. 

Aside from establishing whether \gls{gan} augmentation can lead to an improvement in network performance, we answer the following five important questions:

\noindent\textit{-- Does the choice of segmentation network affect this improvement? \\
-- How does \gls{gan} augmentation compare to rotation augmentation?\\
-- Does the amount of synthetic data added affect this improvement? \\
-- Does the amount of available real data affect this improvement? \\
-- Does the approach generalise to multiple datasets?}

We also explore the distribution of generated images to better understand what modes of augmentation are provided. This allows us to confirm that the \glspl{gan} are producing images which are different to those in the dataset. We show how images are generated with the same pathology, but different unrelated anatomy, and vice versa, demonstrating the ability to perform these particularly challenging forms of augmentation. 

\section{Methods}

We use a \gls{pggan} network~\cite{nvidia} to generate synthetic data. \Gls{pggan} was chosen on the basis of its training stability at large image sizes and apparent robustness to hyperparameter selection.
Whether the choice of \gls{gan} architecture will affect the quality of the augmentation is unclear, however there is evidence~\cite{lucic2017gans} to suggest that different \gls{gan} architectures produce results which are, on average, not significantly different from each other. 

We train a \gls{pggan} on 80k patches sampled from the available training data set as a preprocessing step prior to training a segmentation \gls{cnn}. The \gls{pggan} is trained on multi-channel image patches containing both the acquired image and manual segmentation label, thereby learning the manifold containing this joint data distribution. Synthetic examples are then sampled randomly from this manifold using the trained generator and used to augment the same 80k patches, forming the training data used when training the subsequent segmentation network. The only alteration to the default \gls{pggan} architecture was 
to concatenate a 32x32 layer of Gaussian noise at the start of the fourth (32x32) resolution level when training on \gls{ct} data. This change was found empirically to produce \gls{ct} images with a more realistic noise pattern. The networks were configured to produce images with a size of 128-by-128px with 6 resolution levels.

Segmentation networks were evaluated using training, validation, and test sets.
Performance (measured by \gls{dsc}) on the validation set was monitored during training with the best model at the conclusion of training applied to the test set.

A set of experiments were designed to assess effect of introducing \gls{gan} derived synthetic data to a segmentation task. In these experiments, a number of key variables were modified: 

    \emph{Amount of available real data:} To simulate a situation with limited training data, the amount of training images was artificially reduced by randomly selecting a percentage of the available images, prior to sampling the 80k training patches. 
    We performed experiments with percentage reductions in available data ranging from 10\% to 90\%. Note that this reduction is enforced for both the \gls{gan} and segmentation network training stages, ensuring the \gls{gan} is never exposed to more labelled data than the corresponding segmentation network.
    
    \emph{Amount of additional synthetic data:} To investigate whether the amount of synthetic data added to the real data affects the performance of a segmentation network, experiments were run with different amounts of additional synthetic data. To ensure equal access to the information available in the real data between experiments, synthetic data is added to the real data, increasing the size of the dataset, rather than replacing real data. The amount of additional patches is expressed as a percentage of the real patches. For example an experiment with +50\% synthetic data would use 120k patches (80k real and 40k synthetic).
    
    \emph{Dataset:} Two different datasets are explored to assess the ability for \gls{gan} augmentation to generalise across segmentation tasks. The first dataset contains \gls{ct} images with manually delineated \gls{csf} labels split into into 3 classes: cortical \gls{csf}, brain stem \gls{csf} and ventricular \gls{csf}. Data is split in the same way as in~\cite{Liang}, using the same preprocessing and sampling procedures. This provides 500 manually labelled training image slices, with an additional 282 validation slices, from 101 subjects. For these experiments, the average \gls{dsc} is used as the primary measure of performance, though results across each class are also analysed. The second dataset contains \gls{flair} images with manual \gls{wmh} segmentations. 147 \gls{flair} image stacks were acquired as described in~\cite{Edin2}. These were manually segmented, before being bias corrected, brain extracted, rigidly co-registered and intensity normalised as in~\cite{Huppertz2011}, and randomly split into equal sized training, validation and test sets. By selecting two dissimilar tasks (multi- and single-class segmentation) across two modalities (\gls{ct} and \gls{mr}) we cover a wide range of likely applications for \gls{gan} augmentation.

    \emph{Segmentation network:} We investigate three different segmentation networks across the experiments. In~\cite{Liang}, the authors show that both UNet and Residual UNet (UResNet)~\cite{Ricardo} architectures perform well on this \gls{ct} dataset, we therefore choose to explore both of these. The same hyperparameters were used as in~\cite{Liang}. DeepMedic~\cite{CRF} is a popular general purpose segmentation algorithm which has been shown to perform well in many applications, and was therefore chosen as a third network to explore. DeepMedic was modified only so as to accept 128x128 \gls{2d} patches. Between these three, we represent the most popular \gls{cnn} architectures currently in use.
    
    \emph{Augmentation:} As discussed in Section~\cite{deform} extensive augmentation, beyond simple reflection, is rarely used in brain imaging due to the variety of preprocessing options available and anatomical constraints of the brain. However, in order to examine the interaction of \gls{gan} and rotation augmentation we elect not to perform coregistration on the \gls{ct} dataset. Of the other common forms of augmentation, reflection augmentation is routinely performed in all experiments, translation augmentation is encapsulated in the patch based approach, intensity augmentation is obviated by intensity normalisation, and deformations are not considered due to anatomical constraints (preserving shape, symmetry etc.).

\begin{table}[h]
\centering
\caption{Summary of experiments}
\label{tab:experiments}
\begin{tabular}{@{}p{2.7cm}p{2.3cm}p{2.3cm}p{1.3cm}p{3cm}@{}}
\toprule
\% of available real data sampled from& \% added synthetic data & Segmentation network & Dataset & Augmentation type                 \\ \midrule
100, 50, 10            & 0, 50, 100              & UNet, UResNet       & CT      & Rotation+GAN                              \\
100, 50, 10            & 0, 100                  & UNet                 & CT      & None, GAN, rotation, rotation+GAN            \\
100, 50, 10            & 0, 12.5, 25, 37.5, 50, 100 & UNet                 & CT      & Rotation+GAN                               \\
100, 90...20, 10      & 0, 50                   & UNet                 & CT      & Rotation+GAN                                \\
100, 50, 10            & 0, 50, 100              & DeepMedic            & MR      & GAN                                     \\ \bottomrule
\end{tabular}
\end{table}

\begin{figure}[h]
\centering
\begin{subfigure}{.25\textwidth}
  \centering
  \includegraphics[angle=90,trim={5.3cm 7.47cm 10.25cm 15.9cm},clip,width=0.75\linewidth]{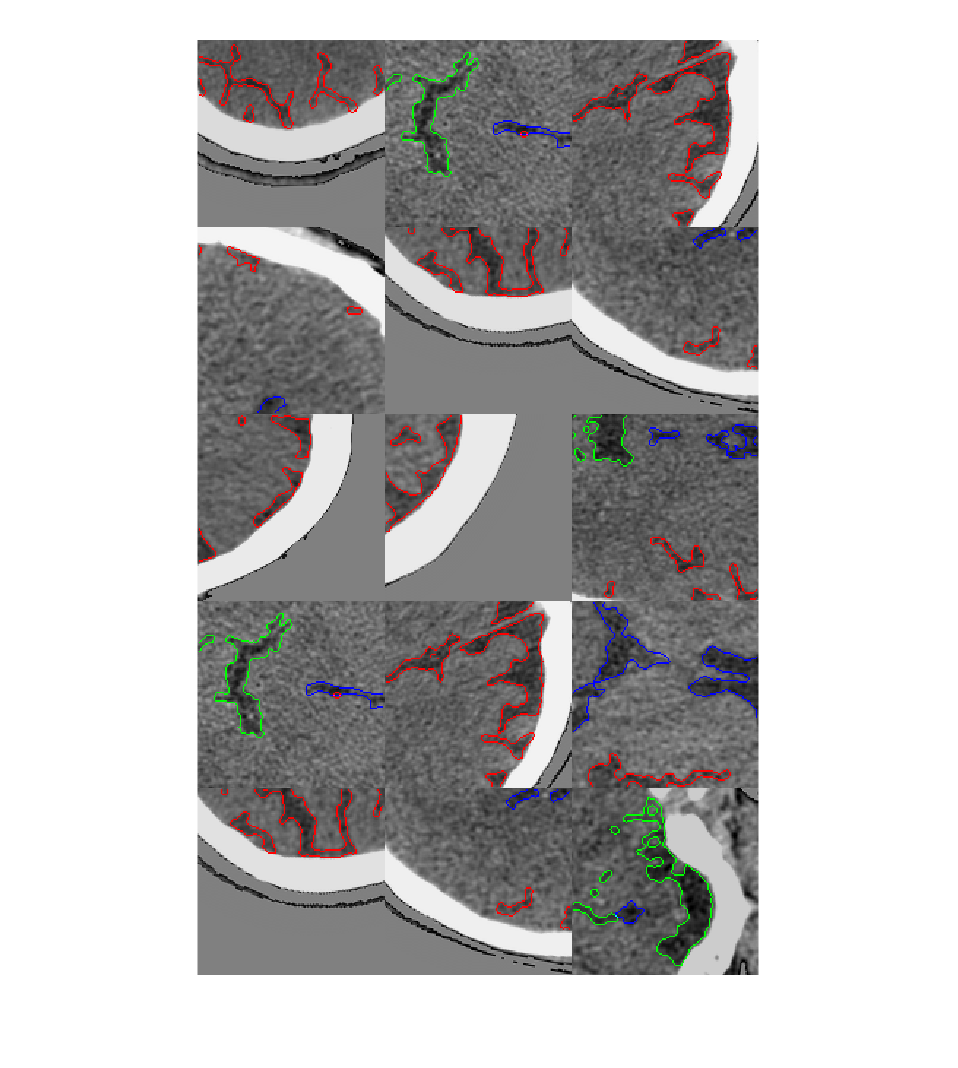}
  \caption{Real CT}  
\end{subfigure}%
\begin{subfigure}{.25\textwidth}
  \centering
  \includegraphics[angle=90,trim={5.3cm 12.37cm 10.25cm 11cm},clip,width=0.75\linewidth]{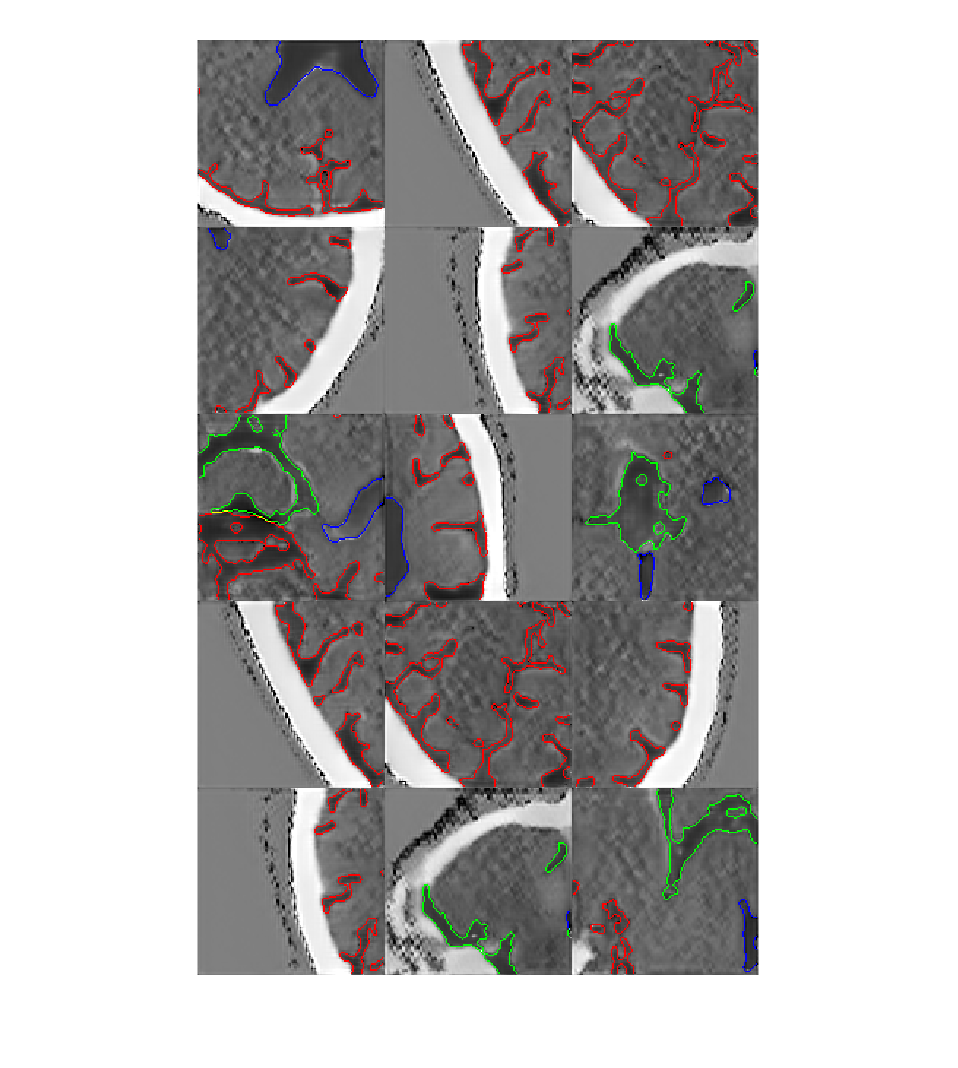}
  \caption{Synthetic CT}
\end{subfigure}%
\begin{subfigure}{.25\textwidth}
  \centering
  \includegraphics[angle=90,trim={6.0cm 1.7cm 2.5cm 0.8cm},clip,width=0.75\linewidth]{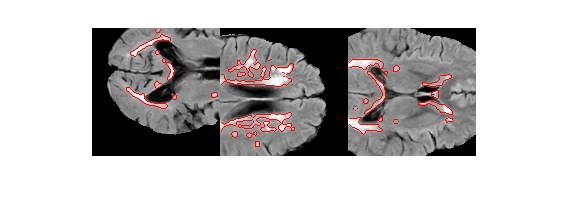}
  \caption{Real MRI}  
\end{subfigure}%
\begin{subfigure}{.25\textwidth}
  \centering
  \includegraphics[angle=90,trim={6.0cm 1.7cm 2.5cm 0.8cm},clip,width=0.75\linewidth]{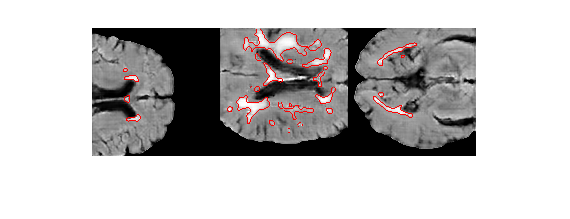}
  \caption{Synthetic MRI}
\end{subfigure}%

\caption{Examples of real and \gls{gan} generated synthetic patches. \textit{Left} \gls{csf}. Red: Cortical \gls{csf}. Green: Brain stem \gls{csf}. Blue: Ventricular \gls{csf}. \textit{Right:} \gls{wmh}.}
\label{fig:Real_synth}
\end{figure}

Table~\ref{tab:experiments} summarises the five sets of experiments which were carried out to answer the questions posed earlier. In each experiment, the segmentation network is treated as a black box and unchanged. This provides a fair platform upon which to observe the effects of \gls{gan} augmentation by ensuring that any changes in performance are as a result of the additional synthetic data, and not of changes in the network itself. \Gls{gan} training took 36 hours, each UNet took 4 hours, each Res-UNet took 24 hours and each DeepMedic network took 24 hours on an Nvidia GTX 1080 Ti or similar GPU. All segmentation experiments on CT were repeated 8 times, while those on MR were repeated 14 times to compensate for a higher observed variance. Examples of real and synthetic patches generated for each dataset can be seen in Figure~\ref{fig:Real_synth}.

\section{Results}
\subsection{Segmentation results}
The following tables and graphs show the results over the two sets of experiments. All tables show the average \gls{dsc}, with the standard deviation in brackets. Results which are statistically different from the baseline (2-tailed t-test, 5\% significance level) are shown in bold.

\begin{center}
\captionof{table}{\textbf{\gls{csf} segmentation on CT:} Results with different proportions of the available training data and varying amounts of additional synthetic data using UNet and UResNet architectures.}
\label{tab:UResNet_UNet}
\small
\begin{tabular}{@{}p{0.57cm}P{0.8cm}P{1.63cm}P{1.72cm}P{1.72cm}P{1.63cm}P{1.72cm}P{1.72cm}@{}}
\toprule
                  &          & \multicolumn{6}{c}{Available data}                                                \\ \midrule
                  &          & \multicolumn{3}{c}{UNet}                & \multicolumn{3}{c}{UResNet}             \\
                  &          & 100\%       & 50\%        & 10\%        & 100\%       & 50\%        & 10\%        \\ \midrule
\multirow{3}{*}{\rotatebox[origin=c]{90}{\parbox[c]{1.45cm}{\centering Additional Data}}} & 0\%   & 88.9 (0.51) & 86.0 (0.50) & 76.9 (0.58) & 86.8 (0.82) & 82.7 (1.55) & 72.5 (1.98) \\ \rule{0pt}{3ex}  
                  & 50\%  & 89.2 (0.30) & \textbf{87.3} (0.46) & \textbf{78.6} (1.04) & 86.3 (1.44) & \textbf{84.3} (1.31) & 74.3 (1.63) \\ \rule{0pt}{3ex}  
                  & 100\% & 89.3 (0.39) &\textbf{86.9} (0.36)& \textbf{78.4} (0.99) & 86.3 (1.24) & 84.1 (1.32) & \textbf{74.7} (1.18) \\ \bottomrule
\end{tabular}
\end{center}

\begin{center}
\captionof{table}{\textbf{\gls{csf} segmentation on CT:} UNet results with different proportions of the available training data and different augmentation techniques.}
\label{tab:aug}
\small
\begin{tabular}{@{}cC{2.4cm}C{2.4cm}C{2.4cm}C{2.4cm}@{}}
\toprule
                                                  & \multicolumn{3}{c}{Available data}      \\ \midrule
                                                  & 100\%       & 50\%        & 10\%        \\ \midrule \rule{0pt}{0ex}
 No augmentation             & 88.1 (0.32)  & 85.0 (0.58)  & 75.1 (0.60) \\ \rule{0pt}{2ex}    
                                  GAN augmentation            & 88.4 (0.41)  & 85.6 (1.33) & 76.3 (1.77) \\ \rule{0pt}{2ex}    
                                  Rotation augmentation           & \textbf{88.9} (0.51) & \textbf{86.0} (0.50) & \textbf{76.9} (0.58) \\ \rule{0pt}{2ex}
                                  GAN + Rotation augmentation           & \textbf{89.3} (0.39) & \textbf{86.9} (0.36) & \textbf{78.4} (0.99) \\                                   \bottomrule
\end{tabular}
\end{center}

\begin{figure}[H]
\includegraphics[trim={2.8cm 0 3cm 0},clip,width=1\linewidth]{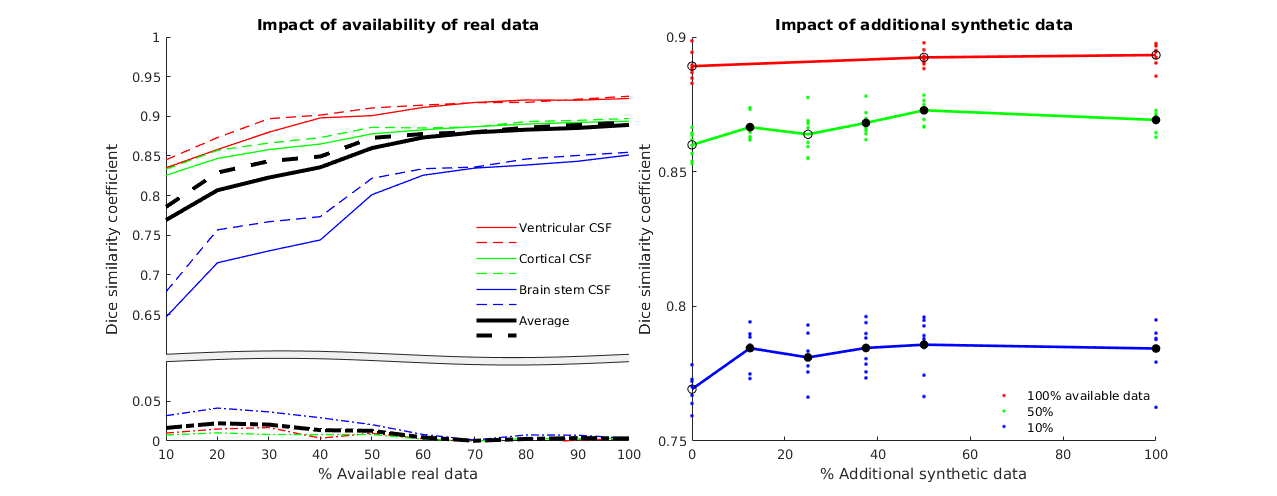}
\captionof{figure}{\textbf{\gls{csf} segmentation on CT:} \textit{Left:} Average \gls{dsc} for each class (coloured) and mean across classes (black) as availability of real data varies. Solid lines show performance without \gls{gan} augmentation, dashed lines show performance with +50\% synthetic data, and dot/dashed lines show the difference, indicating the improvement seen with \gls{gan} augmentation. \textit{Right:} Average \gls{dsc} observed using a UNet as synthetic data is added, when 100\%, 50\% and 10\% of the total amount of real data is used. Each coloured dot represents an experiment. Black circles show the mean with filled circles indicating results significantly different from the baseline.}
\label{fig:Curve_UNet}
\end{figure}

\begin{center}
\captionof{table}{\textbf{\gls{wmh} segmentation on MRI:} Results with different proportions of the available training data and varying amounts of additional synthetic data.}
\label{tab:wmh}
\small
\begin{tabular}{@{}cP{1.8cm}P{3.1cm}P{3.1cm}P{3.1cm}@{}}
\toprule
                                 &                 & \multicolumn{3}{c}{Available data}      \\ \midrule
                                 &                 & 100\%       & 50\%        & 10\%        \\ \midrule
\multirow{3}{*}{\rotatebox[origin=c]{90}{\parbox[c]{1.45cm}{\centering Additional Data}}} & 0\%             & 66.0 (1.26) & 61.4 (2.67) & 52.2 (6.65) \\ \rule{0pt}{3ex}    
                                 & 50\%            & 65.5 (1.21) & \textbf{63.7 (0.69)} & \textbf{57.2 (4.09)} \\ \rule{0pt}{3ex}    

                                 & 100\%           & \textbf{64.8 (1.34)} & 62.8 (1.17) & 55.7 (4.26) \\ \bottomrule
\end{tabular}
\end{center}

\subsection{Qualitative evaluation}

\begin{figure}[h]
\centering
\begin{subfigure}{.27\textwidth}
  \centering
\includegraphics[angle=90,trim={0cm 23.075cm 0cm 4.575cm},clip,width=1\linewidth]{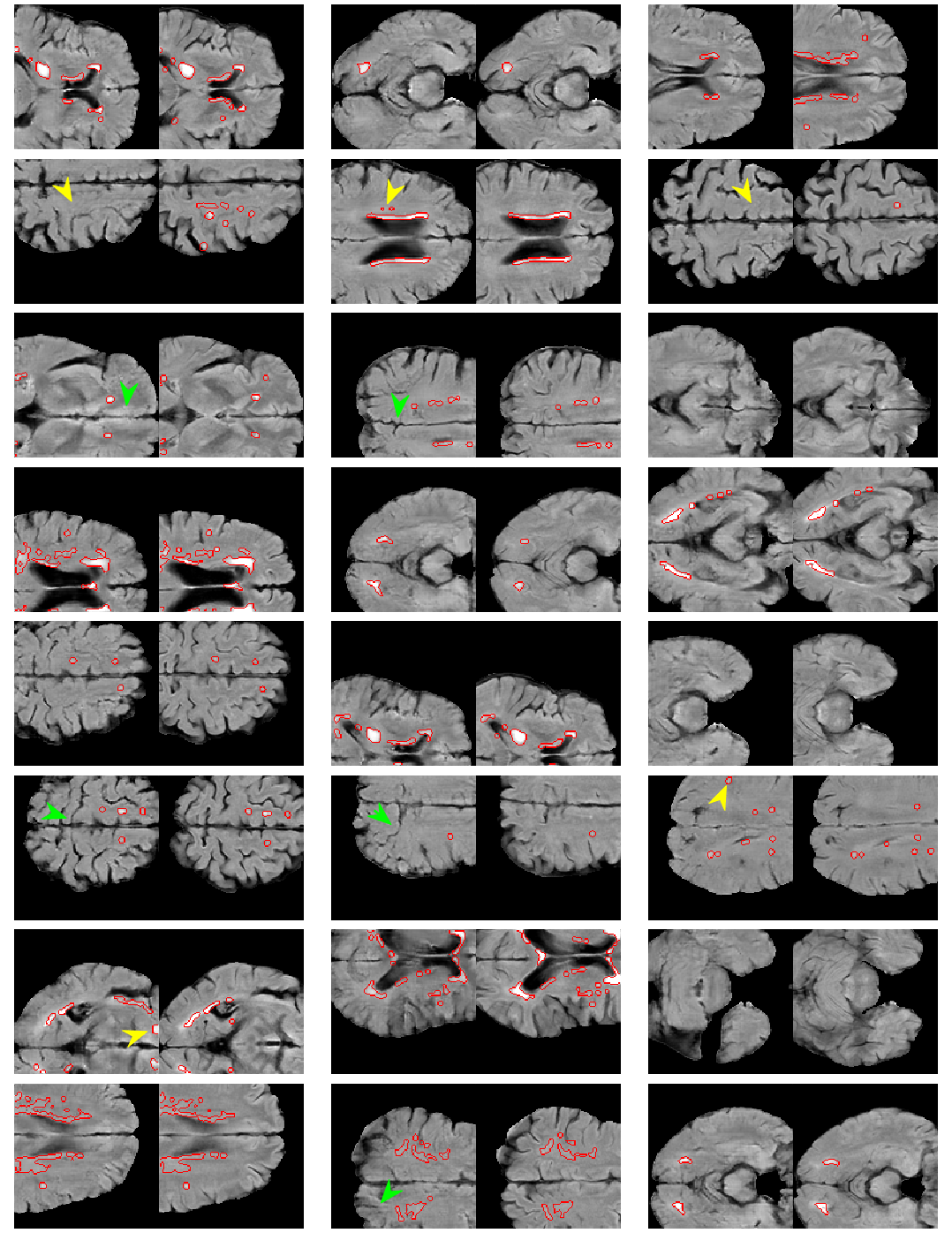}
  \caption{5 training images}
\end{subfigure}~~~~~~%
\begin{subfigure}{.27\textwidth}
  \centering
\includegraphics[angle=90,trim={0cm 27.7cm 0cm 0cm},clip,width=1\linewidth]{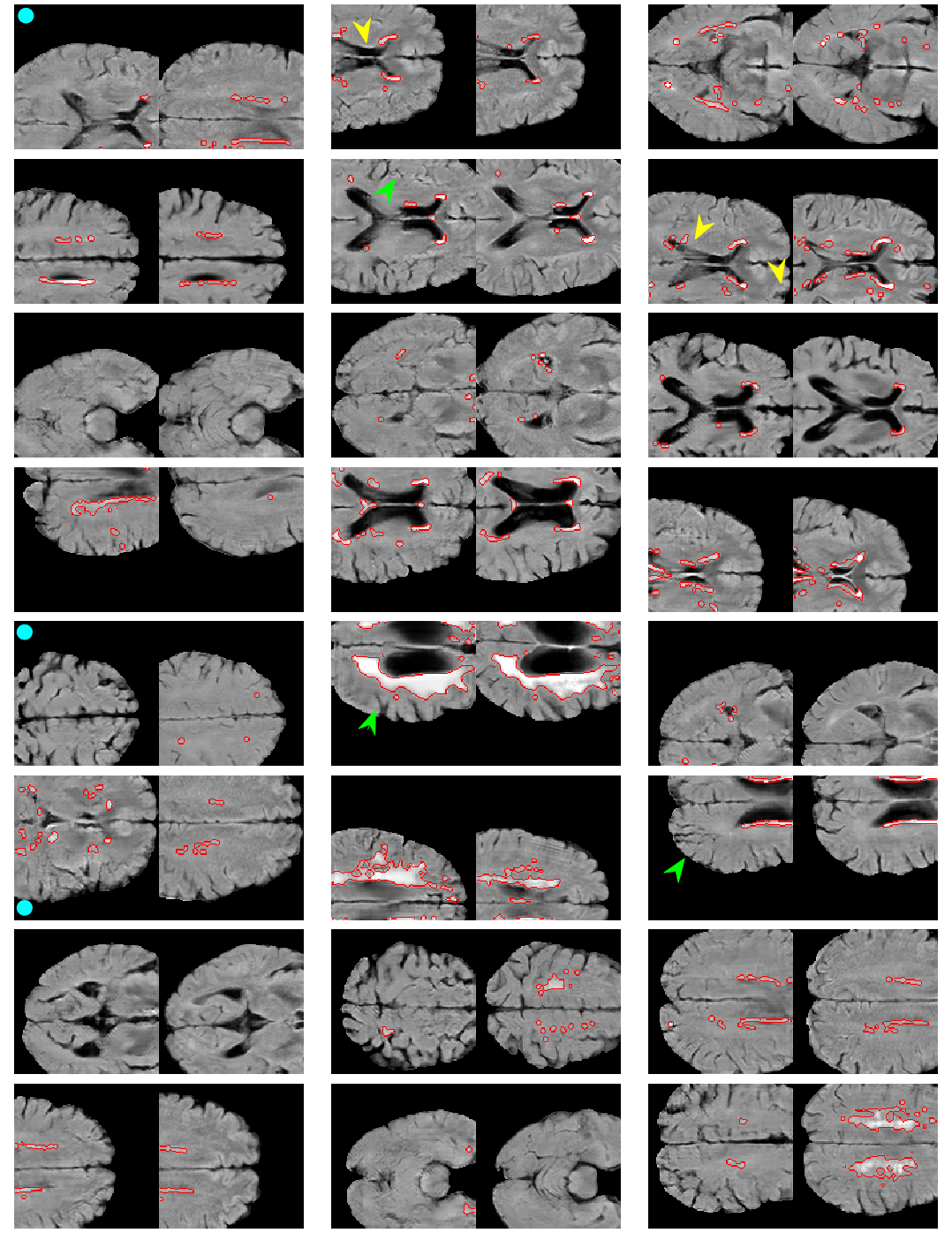}
  \caption{25 training images}
\end{subfigure}~~~~~~%
\begin{subfigure}{.27\textwidth}
  \centering
\includegraphics[angle=90,trim={0cm 13.875cm 0cm 13.875cm},clip,width=1\linewidth]{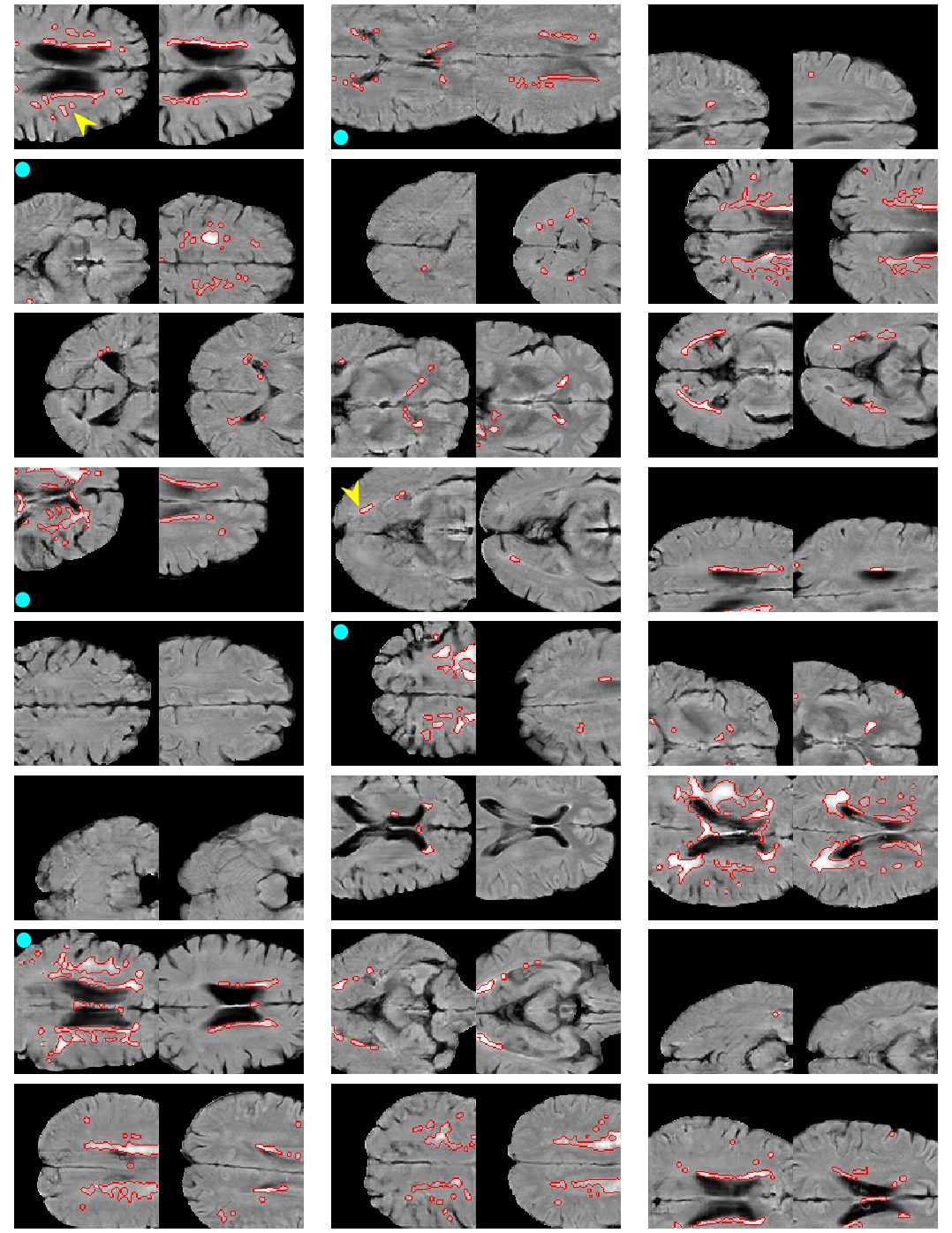}
  \caption{50 training images}  
\end{subfigure}%

\caption{Synthetic images (top of pair) with their nearest neighbours in the training set (bottom of pair) from \glspl{gan} trained on patches from 5, 25 and 50 real \gls{mr} images. Some local signs of successful augmentation are indicated using green (same lesions, different anatomy) and yellow (same anatomy, different lesions) arrows, and novel images (new anatomy and lesions) are shown with blue dots.}
\label{fig:Qual100pc}
\end{figure}

 
As well as the quantitative segmentation results, the generated \gls{mr} images were also compared to their nearest neighbour in the training set to elucidate what extra information \gls{gan} augmentation provides. These images, a subset of which are shown in Figure~\ref{fig:Qual100pc}, were examined looking for cases where: lesions were duplicated on different anatomy; lesions were changed whilst anatomy stays the same; the nearest neighbour is substantially different. The latter indicates the \gls{gan} has learned a smooth manifold leading to potentially novel anatomy.

\section{Discussion}
It can be seen from across all of the results that \gls{gan} augmentation can provide a modest but significant improvement in segmentation performance in many cases. By far the strongest factor controlling the improvement seen is the amount of real data available for training. There is a clear trend across all results that the greatest improvements can be seen in the cases where real data is the most limited. However, Figure~\ref{fig:Curve_UNet} suggests that there is perhaps a drop in improvement seen at the very lowest levels of available data, likely due to there being too little data to properly train the \gls{gan}. Results on the \gls{ct} data suggest that there are no circumstances in which using synthetic data leads to worse results even when large amounts of real data is available. However, this is not reflected in the \gls{mr} results in Table~\ref{tab:wmh}, where a loss in \gls{dsc} is observed when all the data is used. This suggests a tipping point associated with the amount of available data, beyond which \gls{gan} augmentation harms rather than helps. 

Another benefit of \gls{gan} augmentation can be seen in the \gls{dsc} observed on the individual \gls{csf} classes in Figure~\ref{fig:Curve_UNet}. A ratio of 1.35:4.35:1 between ventricular, cortical and brain stem \gls{csf} classes in the training set indicates a moderate class imbalance, with examples of the former and latter being relatively limited. Figure~\ref{fig:Curve_UNet} shows that it is these two classes which benefit most from \gls{gan} augmentation. Brain stem \gls{csf} segmentation appears to benefit the most, though this can be attributed to ventricular \gls{csf} segmentation being an inherently  easier proposition, and therefore being consistently well segmented anyway.

Table~\ref{tab:UResNet_UNet} shows that there is little difference in the effect of \gls{gan} augmentation when using different segmentation networks. This, coupled with the \gls{wmh} results, suggests that \gls{gan} augmentation may benefit any segmentation network, regardless of architecture. Similarly, Figure~\ref{fig:Curve_UNet} shows that a similar level improvement is seen across a broad range of additional synthetic data quantities. 
The amount of synthetic data is therefore not an additional parameter which needs to be finely tuned. This, coupled with the earlier observation that synthetic data rarely impairs performance, makes \gls{gan} augmentation a practical proposition.

It is interesting to note that the improvements given by using both traditional and \gls{gan} augmentation, as seen in Table~\ref{tab:aug}, are consistently more than the sum of the improvements given by using the two methods separately. 
This provides strong evidence that the additional information provided by the two augmentation methods are independent. It also suggests that when used together they are potentially synergistic, an observation which agrees with the results in~\cite{Frid-Adar2018}. This could be due to the two methods acting in different ways, with \glspl{gan} providing an effective alternative to traditional augmentation when attempting to interpolate within the training distribution, but cannot extrapolate beyond its extremes without the aid of traditional augmentation like rotation. 

Figure~\ref{fig:Qual100pc} provides an interesting insight into what additional information is being provided by \gls{gan} augmentation. In the case of 5 training images, it is clear that each generated image is based heavily on an image from the training set. This is unsurprising as there are very few images to train on, and little variation which can be learned. However, there are subtle differences present in the majority of synthetic images. There are cases where lesions present in the real image are not reproduced in the synthetic image, as well as cases where the shape and number of lesions present in the synthetic image differ from those in the real image. Both of these effects can be extremely valuable to prevent overfitting when training a model - the former decoupling the presence of lesions from the surrounding anatomy, and the latter providing more variety of pathology. When the number of training images increases to 25, we begin to see cases where there are no close matches in the training set, in addition to the cases of differing anatomy and pathology seen previously. This trend gets even stronger in Figure~\ref{fig:Qual100pc} where all 50 training images are used. There are often substantial differences between the synthetic images and their closest real image, suggesting that the \gls{gan} has learned to produce data substantially beyond what was provided to it. We also observe that these modifications appear reasonable in all cases, with no obvious unrealistic lesions or anatomy being synthesised.



\subsection{Conclusion}

This paper has investigated augmenting training data using \gls{gan} derived synthetic images, and demonstrated that this can improve results across two segmentation tasks. The approach has been shown to work best in cases of limited data, either through a lack of real data or as a result of class imbalance. 
\Gls{gan} augmentation requires little overhead, involving only the training of a single out-of-the-box \gls{gan}, does not involve optimising additional parameters, and has been shown to be low-risk by not hurting performance when training data is limited. 
A conservative interpretation of the results from the typical tasks explored here suggests that in cases where $5-50$ labelled image volumes are available, augmenting these with an additional $10-100\%$ \gls{gan} derived synthetic patches has the potential to lead to significant improvements in DSC. 

One major advantage that traditional augmentation has over \gls{gan} augmentation is the ability to extrapolate. \Glspl{gan} can provide an effective way to fill in gaps in the discrete training data distribution and augment sources of variance which are difficult to augment in other ways, but will not extend the distribution beyond the extremes of the training data. In general, appropriate traditional augmentation procedures should be used to extrapolate and extend the manifold of semantically viable images. \Glspl{gan} can then be used to interpolate between the discrete points on this manifold, providing an additional data driven source of augmentation. Future work will involve investigating \gls{gan} augmentation in other areas, and to evaluate the impact of different \gls{gan} architectures.

\bibliographystyle{splncs04.bst}
\bibliography{bib.bib}

\end{document}